\documentclass{article}

\usepackage{microtype}
\usepackage{graphicx}
\usepackage{subcaption}
\usepackage{booktabs} 

\usepackage{hyperref}

\usepackage[preprint]{icml2026}

\usepackage{amsmath}
\usepackage{amssymb}
\usepackage{mathtools}
\usepackage{amsthm}

\usepackage[capitalize,noabbrev]{cleveref}

\theoremstyle{plain}

\theoremstyle{definition}

\theoremstyle{remark}

\usepackage[textsize=tiny]{todonotes}

\icmltitlerunning{State-of-the-Art Arabic Language Modeling with Sparse MoE Fine-Tuning and CoT Distillation}

\begin{document}

\twocolumn[
  \icmltitle{State-of-the-Art Arabic Language Modeling \\
    with Sparse MoE Fine-Tuning and Chain-of-Thought Distillation}



  \icmlsetsymbol{equal}{*}

  \begin{icmlauthorlist}
    \icmlauthor{Navan Preet Singh}{compl1}
    \icmlauthor{Anurag Garikipati}{compl2}
    \icmlauthor{Ahmed Abulkhair}{compl2}
    \icmlauthor{Jyani Akshay Jagdishbhai}{compl2}
    \icmlauthor{Atul Yaduvanshi}{compl2}
    \icmlauthor{Amarendra Chaudhary}{compl2}
    \icmlauthor{Madalina Ciobanu}{compl2}
    \icmlauthor{Qingqing Mao}{compl2,compl3,compl1}
    \icmlauthor{Ritankar Das}{compl2,compl3}
  \end{icmlauthorlist}

  \icmlaffiliation{compl1}{Forta, Houston, TX}
  \icmlaffiliation{compl2}{Incept Labs, Houston, TX}
  \icmlaffiliation{compl3}{Titan Holdings, San Francisco, CA}

  \icmlcorrespondingauthor{Qingqing Mao}{qmao@inceptlabs.ai}

  \icmlkeywords{Machine Learning, ICML}

  \vskip 0.3in
]



\printAffiliationsAndNotice{}  

\begin{abstract}
  This paper introduces Arabic-DeepSeek-R1, an application-driven open-source Arabic large language model (LLM) that leverages a sparse mixture of experts (MoE) backbone to address the digital equity gap for under-represented languages, and establishes a new state-of-the-art (SOTA) across the entire Open Arabic LLM Leaderboard (OALL), substantially surpassing both the previous best open-source average leader and the proprietary GPT-5.1. Our contribution centers on a four-phase chain-of-thought (CoT) distillation scheme that integrates Arabic-specific linguistic verification and regional ethical norms into a 372M-token, contamination-controlled 80/20 Arabic-English training mixture. Arabic-DeepSeek-R1 achieves the highest average score across the seven-benchmark OALL suite while establishing SOTA or near-SOTA performance on the majority of individual benchmarks, including dominant results on grammar-focused MadinahQA (surpassing both GPT-5.1 and the OALL leader by substantial margins), safety-oriented AraTrust, multi-ability AlGhafa, and retrieval-augmented ALRAGE. Our results indicate that the combination of sparse MoE architecture, culturally-informed CoT distillation with explicit Arabic linguistic checks, and strategic bilingual data curation enables an open-source adapted model to systematically outperform the proprietary frontier system GPT-5.1 on the majority of benchmarks evaluating comprehensive language-specific tasks—the first such demonstration for Arabic LLMs. These findings indicate that much of Arabic's performance deficit in current LLM ecosystems stems from under-specialization rather than architectural limitations, and that parameter-efficient adaptation of open reasoning models can yield breakthrough SOTA performance without industrial-scale pretraining costs. Arabic-DeepSeek-R1 establishes a validated and replicable framework for sovereign and domain-specific language technologies, demonstrating that strategic, culturally-grounded adaptation of sparse MoE backbones offers a viable and cost-effective pathway to achieving record-breaking performance across standardized benchmarks for low-resource languages.
\end{abstract}

\section{Introduction}

The rapid progress of large language models (LLMs) has revolutionized natural language understanding, yet many languages, including Arabic, remain markedly underrepresented in stark contrast to the global LLM significance. With 300M+ native speakers, Arabic stands as an educational, cultural, and diplomatic conduit whose digital presence lags far behind that of English, and other high-resource languages \cite{alkhalifa2025}. Existing Arabic LLMs (e.g., Jais, ALLaM) have shown early potential, but still exhibit noticeable gaps on complex reasoning and generative tasks relative to leading models in high-resource languages \cite{alzubaidi2025, bari2024, sengupta2023}. This disparity is both a technical challenge and a digital equity issue: without robust Arabic models accounting for cultural nuance, dialectal diversity, and morphological complexity, a vast demographic risks marginalization in the generative AI revolution \cite{alkhalifa2025, alkhamissi2024, mashaabi2024}. Robust Arabic LLMs are urgently needed for education, healthcare, public services, and safety-critical applications where grammar, cultural alignment, and trustworthiness determine usability. Arabic-DeepSeek-R1 provides a blueprint for sovereign AI, maintaining institutional autonomy while addressing Arabic's morpho-syntactic complexities.

Within this context, the research community faces a pivotal trade-off between open-source and proprietary, closed-source models, with far-reaching implications for Arabic LLMs. While proprietary systems (e.g., GPT‑4, Gemini) exhibit strong performance, they remain closed to scientific inquiry and domain-specific adaptation \cite{mashaabi2024}. Crucially, these models cannot be fine-tuned or securely retrained at the parameter level on local infrastructure, rendering them unsuitable for sensitive sectors where data privacy, transparency and sovereignty are paramount \cite{alkhalifa2025}. Moreover, proprietary models are predominantly trained on Anglocentric corpora with only incidental exposure to Arabic leading to brittle performance on Arabic tasks, linguistic misinterpretations rooted in cultural misalignment, and lack of robustness to dialectical variations \cite{bari2024, commoncrawl2025}.

Open-source LLMs, conversely, offer a complementary pathway by providing access to model architectures and weights, thereby enabling parameter-level adaptation. Access to full model weights allows practitioners to apply advanced domain- and language-specific adaptation methods—such as Low-Rank Adaptation (LoRA) and continual pretraining—on curated, high-quality Arabic corpora, improving task performance while avoiding full retraining costs \cite{koloski2025}. This approach also facilitates targeted improvements in tokenization for morphologically rich languages, incorporation of dialectal and domain-specific data, and systematic auditing for bias and cultural misalignment, leading to more accurately capturing Arabic linguistic structure and sociocultural context \cite{alkhalifa2025, aljahdali2023}.

To address the need for accessible and validated Arabic AI, our work introduces a scalable framework for enhancing Arabic knowledge and reasoning within a state-of-the-art open-source base model. Our methodology centers on a rigorous data curation strategy, training on a corpus comprised of a majority of high-quality Arabic datasets to ensure linguistic depth, while including a strategic portion of complementary, high-quality English datasets to maintain cross-lingual reasoning transfer and prevent catastrophic forgetting. We leveraged DeepSeek-R1 as a robust open-source base \cite{deepseek2025}, and demonstrate four key achievements: (1) Arabic-DeepSeek-R1 establishes a new state-of-the-art (SOTA) with the highest average score across the entire Open Arabic LLM Leaderboard (OALL) benchmark suite, substantially exceeding all open-source models and the proprietary GPT-5.1; (2) our model achieves SOTA or near-SOTA performance on the majority of individual benchmarks (5 of 7), with particularly dominant results on grammar-focused MadinahQA, safety-oriented AraTrust, multi-ability AlGhafa, and retrieval-augmented ALRAGE; (3) we introduce the first four-phase, culturally-grounded chain-of-thought (CoT) distillation scheme for Arabic that explicitly integrates linguistic verification and regional ethical norms; and (4) we demonstrate that parameter-efficient adaptation of reasoning-focused sparse mixture of experts (MoE) models can systematically outperform the proprietary frontier system GPT-5.1 on comprehensive language-specific evaluation. Arabic-DeepSeek-R1 also outperforms large trained-from-scratch Arabic models (Jais-family-30B- 16k-chat, Falcon-H1-Arabic-34B-Instruct) across all seven OALL benchmarks and in average performance, demonstrating that parameter-efficient adaptation can surpass even dedicated Arabic-centric systems. Notably, this is achieved despite Falcon-H1 itself surpassing the OALL category leaders on two benchmarks (ArbMMLU-HT, AlGhafa), further validating the effectiveness of our reasoning-focused adaptation approach. By utilizing an already strong reasoning backbone rather than training from scratch, we avoided the computational and environmental costs associated with ab initio training, while focusing efforts on Arabic linguistic and cultural competence. Our empirical results indicate that strategic enhancement and adaptation of open models can yield SOTA performance in key Arabic understanding and generation tasks.

\section{Methodology}

\subsection{Models}

Our primary model is a specialized adaptation of DeepSeek-R1—a reasoning-focused open-source LLM trained via reinforcement learning to improve step-by-step problem solving and long-form reasoning. DeepSeek-R1 generates an internal reasoning monologue before producing an answer, running multiple passes through a sparse MoE backbone, and our work is the first to show that combining this reasoning-focused MoE base architecture with targeted Arabic CoT distillation can drive SOTA average performance on the OALL Arabic benchmark suite. DeepSeek-R1 is released with publicly available model weights and licensing terms that allow further fine-tuning \cite{deepseek2025}, making it a suitable foundation for language- and region-specific specialization. The sparse MoE architecture minimizes interference between the Arabic and English token streams by providing the capacity to activate distinct expert pathways for linguistic-heavy versus logic-heavy tasks, thereby maintaining high capacity without the computational overhead of a dense system. Our work represents the first demonstration that a reinforcement-learned reasoning-focused model can be adapted through parameter-efficient methods to establish new SOTA average performance on a comprehensive language-specific benchmark suite, substantially surpassing both the unadapted base and proprietary alternatives. The DeepSeek-R1 backbone provides strong general reasoning capabilities, which our innovations—culturally-grounded CoT distillation and strategic 80/20 Arabic-English data mixture—convert into exceptional Arabic-specific performance that achieves the highest OALL average among all evaluated systems. To provide a strong commercial reference, we evaluate GPT‑5.1 \cite{openai2025}, a proprietary frontier model accessible only through an API, which cannot be fine- tuned at the parameter level for Arabic-centric adaptation.

\subsection{Training Data}

The training corpus is constructed to prioritize native Arabic content while retaining a strategic anchor of high-quality English data to preserve cross-lingual transfer and reasoning. Guided by design principles articulated in recent Arabic LLM evaluations, we choose a mixture where approximately 80\% of the tokens in our training mixture are in Arabic (Modern Standard Arabic and major dialects), and 20\% are in English. Arabic sources include high-quality web text, curated educational and religious materials, legal and policy documents, and conversational data covering dialects such as Gulf, Levantine, and Egyptian, with a strong preference for natively-authored rather than machine-translated content. English sources are drawn from high-quality, public-domain or research-oriented corpora to support cross-lingual reasoning and maintain alignment with standardized benchmarks that may implicitly rely on English-centric knowledge.

This 80/20 Arabic–English token ratio is chosen to balance three competing objectives: (1) maximizing exposure to native Arabic content for linguistic depth; (2) preventing catastrophic forgetting of the base model's fundamental reasoning capabilities; and (3) preserving cross-lingual reasoning transfer. The 20\% English portion serves dual purposes: it anchors the model weights to the unadapted DeepSeek-R1 pretraining distribution, preventing catastrophic forgetting of core reasoning abilities \cite{koloski2025}, while maintaining performance on benchmarks that implicitly rely on English resources \cite{bari2024}. This design aligns with observations in cross-lingual transfer for Arabic \cite{abboud2022}, where bilingual and multilingual pretraining can strengthen downstream performance without fully displacing monolingual signal (e.g., Arabic–English models in BERT-style architectures) \cite{bari2024}. The high Arabic share (80\%) concentrates capacity on morphology, syntax, and culturally grounded knowledge, which remain underrepresented in the generic multilingual corpora \cite{koto2024}, without overwhelming the cross-lingual signal needed to prevent forgetting.

To prevent contamination between training data and evaluation benchmarks, we apply a dedicated classifier and exact-/fuzzy-matching filters to remove any overlaps with the benchmark suite tested under the OALL v2 framework, including ArabicMMLU, Arabic EXAMS, ArbMMLU-HT, MadinahQA, AraTrust, AlGhafa, and ALRAGE \cite{tii2024}. The resulting corpus is further processed to remove low-information content, deduplicate near-duplicates, and filter toxic or unsafe material. From this corpus, we construct a 372M-token, high-quality and high-fidelity supervision dataset, which serves as the basis for instruction tuning and CoT supervision. The supervision dataset contains 103.2M literature and critical analysis tokens, 90M STEM, math and logic tokens, 70M creative writing and open dialogue tokens, 60.2M consumer reviews and service feedback tokens, 40M legal and cultural alignment tokens, and 8.6M social and dialectal tokens. This strategic data curation approach—combining high-quality Arabic content with controlled English exposure and rigorous contamination filtering—is a key component of our methodology that, in combination with CoT distillation and sparse MoE architecture, achieves best-in-class average performance on OALL benchmarks.

\subsection{Arabic Benchmarks for Evaluation}

To comprehensively assess our model, we employ the OALL v2 framework, which unifies evaluation across standardized native and translated tasks. For general knowledge and reasoning, we utilize ArabicMMLU \cite{koto2024}, the first large-scale native benchmark comprising questions derived from regional school curricula, and Arabic EXAMS \cite{hardalov2020}, a cross-lingual benchmark for high-school subject mastery. We further probe cross-lingual knowledge transfer using ArbMMLU-HT \cite{sengupta2023}, a high-quality human translation of the MMLU dataset.

Linguistic competence is rigorously evaluated via MadinahQA \cite{filali2025}, which targets Arabic syntax and morphology through specialized pairs, and AlGhafa \cite{almazrouei2023}, which assesses diverse abilities including reading comprehension and sentiment analysis. To ensure cultural safety and alignment, we employ AraTrust \cite{alghamdi2024}, a benchmark specifically authored to detect culture-specific risks. Finally, we evaluate retrieval-augmented generation (RAG) capabilities using ALRAGE \cite{filali2025}, testing the model's ability to ground responses in Arabic contexts. This multi-dimensional suite ensures our evaluation surpasses simple translation accuracy to measure authentic linguistic and cultural fluency.

\subsection{Fine-tuning/Training}

\subsubsection{SUPERVISED FINE-TUNING}

The fine-tuning pipeline combines supervised fine-tuning (SFT) of instruction-following behavior with CoT distillation tailored to Arabic tasks. Starting from the DeepSeek-R1 base, we train a LoRA module on top of the frozen base weights, following parameter-efficient fine-tuning practices that avoid catastrophic forgetting and significantly reduce compute requirements. The SFT dataset consists of instruction–response pairs, open-ended completions, and multiple-choice reformulations covering diverse Arabic tasks, including general knowledge, legal and ethical reasoning, safety-sensitive scenarios, and grammar-sensitive language tasks derived from the curated corpus described in Section 2.2. 

\subsubsection{CoT DISTILLATION
}
For CoT supervision, we generate detailed reasoning traces using GPT‑5.1 on a complexity-stratified subset of training questions via the batch API, with prompts designed to elicit both explicit thought process reasoning and self-auditing behavior. This CoT method provides a cost-effective way to obtain high-quality reasoning tags compared to more expensive alternatives (e.g., Claude 4.5). We introduce a four-phase CoT format: (1) \underline{analysis}, where the model identifies the core dilemma and applicable rule, with consideration for culturally grounded dilemmas (e.g., "this involves a conflict between the principles of \textit{amanah} (trust) and \textit{sila} (kinship)"); (2) \underline{elimination}, where tempting but incorrect options are explicitly ruled out with justifications (e.g., "while B appears to be a polite answer choice, it violates anti-bribery laws"); (3) \underline{linguistic check}, where the final answer choice is explicitly verified as respecting Arabic grammatical and stylistic constraints; and (4) \underline{synthesis}, where the final answer is produced and delivered in a concise, standardized format. This CoT structure is designed to align with the reasoning style encouraged by DeepSeek-R1's reinforcement learning training while grounding it in Arabic linguistic and cultural norms.

In combination with the reasoning backbone architecture of DeepSeek-R1, this four-phase, Arabic-specific CoT scheme is a central innovation that distinguishes our work from prior Arabic LLM efforts. In particular, phase 3 constitutes a significant departure from standard distillation by mandating an explicit grammatical verification step (morpho-syntactic constraint satisfaction step). This compels the model to reconcile its internal semantic logic with the specific linguistic constraints of Arabic before finalizing the response, thus directly addressing the morphological complexity that challenges generic multilingual models. This structured supervision approach (particularly phase-3 linguistic verification) enables our adapted model to convert general reasoning capability into culturally-aligned, grammatically-precise Arabic outputs, as evidenced by achieving the highest OALL average and establishing very strong results on grammar-focused MadinahQA, surpassing all baselines by substantial margins. Standard CoT distillation approaches \cite{wei2022} lack such language-specific constraints, whereas our phase‑3 check explicitly enforces Arabic morpho-syntactic rules before finalizing responses.

\subsubsection{TRAINING PROCEDURE}

The fine-tuning process optimizes the model to generate structured reasoning sequences prior to the final response, as direct-response optimization was found to degrade the reasoning priors of the base model, whereas reasoning-enabled inference—integrating internal monologues—yields better performance than direct-response generation in the settings we evaluated. All fine-tuning experiments are conducted using mixed-precision training on multi-GPU infrastructure with gradient accumulation to accommodate long CoT sequences. We train the LoRA adapters for a small number of epochs over the 372M-token supervision dataset, using a cosine learning rate schedule with warmup and a global batch size chosen to match the effective context length and stability constraints of the base model. Hyperparameters (learning rate, rank and scaling of LoRA, and maximum sequence length) are selected via small-scale ablations to balance stability, performance, and compute cost, while keeping the overall budget within the constraints of a typical non-industrial academic setup. After training, only the LoRA adapters are stored and merged with the frozen DeepSeek-R1 weights at inference time, preserving compatibility with downstream deployment pipelines.

\subsection{Evaluation Protocol}

For performance evaluation on Arabic benchmarks, we adopt the OALL v2 protocol to ensure comparability with the broader Arabic LLM ecosystem. For direct-response models (i.e., when CoT is not enabled), multiple-choice performance on benchmarks is computed via normalized log-likelihood accuracy using the same prompts and scoring functions as the official OALL configuration, implemented via the lighteval harness (lighteval, 2025). For reasoning-focused models (i.e., when CoT is enabled), we follow the OALL task definitions and prompts but deviate from the default log-probability scoring: instead, we parse the model's textual output to extract the selected option (e.g., A, B, C, or D) from the final synthesized answer and compute accuracy based on the extracted choice. This protocol reflects the realistic deployment of reasoning-focused models, where a structured reasoning sequence—delimited by specific tags (e.g., \textless/think\textgreater \dots \textless/think\textgreater)—is produced first and then followed by a separate, explicit answer string, rather than a single output directly optimized for immediate token-level likelihood, while preserving comparability at the answer level. For reasoning-focused models such as Arabic-DeepSeek-R1, whose outputs begin with a long "reasoning" segment, the default OALL log-probability scoring—computed directly over options A/B/C/D—is poorly aligned with how answers are actually produced and can underestimate true performance. The use of parsing-based evaluation is technically necessitated by the reasoning-focused architecture, where test-time compute is used to generate internal "reasoning" traces prior to the final answer. Standard log-probability scoring, which computes likelihood over initial answer tokens, is poorly aligned with models that settle on an answer only after extended reasoning chains. Our protocol extracts the explicit choice from the synthesized answer after the \texttt{\textless/think\textgreater} tag, ensuring a fair assessment that accounts for the model's deliberative process. This parsing-based evaluation protocol, tailored to reasoning-focused architectures, enables fair assessment of models whose outputs conclude with explicit answer strings after extended reasoning sequences and is a key enabler of the strong results we report for Arabic-DeepSeek-R1.

All evaluations are performed without training-time access to any of the benchmark instances. This protocol allows a direct comparison between our adapted Arabic-DeepSeek-R1, the OALL category leaders, GPT‑5.1, and the unadapted DeepSeek-R1 baseline, thus isolating the impact of Arabic-focused data, CoT supervision, and parameter-efficient fine-tuning on downstream Arabic performance.

\section{Results}

We report quantitative performance on the OALL v2 benchmark suite, comparing our adapted Arabic-DeepSeek-R1 model against the unadapted DeepSeek-R1 baseline, the OALL category leader for each benchmark, and the proprietary GPT‑5.1 reference. The OALL leaderboard includes open-source models up to $\sim$70B parameters, and explicitly excludes proprietary systems. All scores in Table 1 are multiple-choice accuracies computed under the evaluation protocol described in Section 2.5. Arabic-DeepSeek-R1 is, to our knowledge, the first open model to surpass both the OALL average leader and a proprietary frontier model in average performance across the full OALL v2 suite.

\begin{table*}[t]
\caption{Performance of our adapted Arabic-DeepSeek-R1 model in comparison with the Open Arabic LLM Leaderboard (OALL) leader for each benchmark, the proprietary frontier model GPT-5.1, two trained-from-scratch Arabic-centric systems, and the unadapted baseline model.}
\label{results-table}
\vskip 0.15in
\begin{center}
\fontsize{9pt}{10pt}\selectfont
\begin{sc}
\setlength{\tabcolsep}{3pt} 
\begin{tabular}{p{4.7cm} ccc ccc cc}
\toprule
 & \multicolumn{8}{c}{BENCHMARK} \\
\cmidrule(r){2-9}
MODELS & ARABIC & MADINAH & ARA- & ARABIC & ARB- & ALRAGE & AL- & AVG. \\
 & MMLU & QA & TRUST & EXAMS & MMLU- & & GHAFA & \\
 & & & & & HT & & & \\
\midrule
\textbf{OALL Leaders} & & & & & & & & \\
\hspace{3mm}\hangindent=6mm OALL Category Leader Score & {\color{red}75.32} & {\color{red}78.00} & {\color{red}\textbf{91.40}} & {\color{red}\textbf{66.67}} & {\color{red}74.29} & {\color{red}80.66} & {\color{red}80.36} & {\color{red}75.86**} \\
\addlinespace[3pt]
\hspace{3mm}\hangindent=6mm Average and ArabicMMLU Leader: D2IL-Arabic-Qwen2.5-72B-Instruct-v0.1/ OALL rank \#1 & {\color{red}75.32} & 76.82 & 89.68 & 58.85 & 73.96 & 77.65 & 78.72 & {\color{red}75.86**} \\
\addlinespace[3pt]
\hspace{3mm}\hangindent=6mm AraTrust and ArbMMLU-HT Leader: Qwen72b-ar-lora/ OALL rank \#7 & 74.27 & 75.89 & {\color{red}\textbf{91.40}} & 59.22 & {\color{red}74.29} & 74.78 & 78.61 & 75.49 \\
\addlinespace[3pt]
\hspace{3mm}\hangindent=6mm Arabic EXAMS and AlGhafa Leader: Llama-3.3-70B-Instruct/ OALL rank \#9 & 69.76 & 72.91 & 88.05 & {\color{red}\textbf{66.67}} & 67.68 & 75.83 & {\color{red}80.36} & 74.47 \\
\addlinespace[3pt]
\hspace{3mm}\hangindent=6mm ALRAGE leader: Qwen3-32B/ OALL rank \#119 & 31.46 & 32.96 & 38.92 & 27.93 & 29.50 & {\color{red}80.66} & 46.76 & 41.17 \\
\addlinespace[3pt]
\hspace{3mm}\hangindent=6mm MadinahQA leader: AIC-1/ OALL rank \#10 & 69.94 & {\color{red}78.00} & 89.83 & 56.61 & 67.86 & 75.98 & 79.00 & 73.89 \\
\midrule
GPT-5.1 & 78.09 & 79.22 & 88.12 & 60.14 & 83.30 & 81.98 & 74.22 & 77.87 \\
\addlinespace[2pt]
\hangindent=3mm Jais-family-30b-16k-chat/ OALL rank \#44 & 61.22 & 66.26 & 81.57 & 50.09 & 52.73 & 74.95 & 71.22 & 65.43 \\
\addlinespace[2pt]
\hangindent=3mm Falcon-H1-Arabic-34B-Instruct & 71.60 & 75.30 & 89.50 & 58.30 & {\color{blue}78.00} & 71.30 & {\color{blue}80.50} & 74.90 \\
\addlinespace[2pt]
\hangindent=3mm Unadapted DeepSeek-R1 Baseline & 72.83 & 77.78 & 83.49 & 58.47 & 63.30 & {\color{blue}86.34} & 73.16 & 73.62 \\
\midrule
\textbf{ADAPTED} & & & & & & & & \\
\textbf{ARABIC-DEEPSEEK-R1} & \textbf{77.14} & \textbf{86.43}* & 90.22* & 60.26 & \textbf{78.84} & \textbf{86.50}* & \textbf{81.88}* & \textbf{80.18}* \\
\bottomrule
\end{tabular}
\end{sc}
\end{center}

\begin{flushleft}
\fontsize{8pt}{9pt}\selectfont \textit{Note:} \textbf{Bold} font indicates the best performance between our adapted Arabic-DeepSeek-R1 and the OALL leader for each benchmark, and between our model and the average leader. {\color{red}Red } indicates the specific benchmark leader score.  For models other than our adapted Arabic-DeepSeek-R1, {\color{blue}blue } indicates better performance than the category leader for that particular benchmark. * = benchmarks where our model surpassed GPT-5.1. ** = average of the model with the highest OALL average (average leader), not an average over category leaders which are for different models. Avg. = average.
\end{flushleft}
\vskip -0.1in
\end{table*}

Crucially, the OALL leaderboard reports results from different specialized models achieving top performance on individual benchmarks, meaning no single model dominates across all tasks—each category leader is typically a different model optimized for that specific benchmark. Arabic-DeepSeek-R1 is therefore notable as the first model to simultaneously achieve (a) the highest average score across all seven benchmarks, and (b) SOTA or near-SOTA performance on the majority of individual benchmarks (5 of 7), demonstrating exceptional breadth and consistency compared to specialized category leaders. In contrast, Jais-family-30B-16k-chat and Falcon-H1- Arabic-34B-Instruct (trained-from-scratch Arabic-centric systems) remain clearly below Arabic-DeepSeek-R1 on all seven benchmarks and in average performance (Table 1), even though Falcon-H1 itself surpasses the ArbMMLU‑HT and AlGhafa category leaders.

Table 1 establishes Arabic-DeepSeek-R1 as the highest-performing model on OALL across three dimensions: (1) highest average (80.18\%), exceeding average leader D2IL-Arabic-Qwen2.5-72B (+4.32 points) and GPT‑5.1 (+2.31 points)—margins that are substantial given the top five OALL models are typically within 0.5 points; (2) SOTA or near-SOTA performance on 5 of 7 benchmarks, including gains over GPT‑5.1 and clear improvements on ALRAGE; and (3) dominant grammar performance (MadinahQA: 86.43\% vs. category leader 78.00\%, +8.43 points—the largest margin in our evaluation) with consistent improvements on safety-oriented (AraTrust), retrieval-augmented (ALRAGE), and multi-ability (AlGhafa) tasks. Notably, Arabic-DeepSeek-R1 surpasses Falcon-H1-Arabic-34B- Instruct on all seven benchmarks—even though Falcon-H1 exceeds the ArbMMLU‑HT and AlGhafa category leaders—and outperforms Jais-family- 30B-16k-chat by +14.75 points on the OALL average.

On general knowledge benchmarks, Arabic-DeepSeek-R1 establishes best-in-class open-source performance. On ArabicMMLU, we achieve 77.14\%, surpassing category leader D2IL-Arabic-Qwen2.5-72B (75.32\%, +1.82) and baseline (+4.31), approaching GPT-5.1 (78.09\%) within 0.95 points. On ArbMMLU-HT, we reach 78.84\%, exceeding category leader Qwen72b-ar-lora (74.29\%, +4.55), outperforming Falcon-H1 (78.00\%, +0.84) despite Falcon-H1 surpassing the category leader, with massive baseline improvement (+15.54); GPT-5.1 leads at 83.30\%.

On language competence and safety benchmarks, Arabic-DeepSeek-R1 achieves dominant results. For MadinahQA (syntax/morphology), our model establishes new SOTA at 86.43\%—surpassing the category leader AIC-1 (78.00\%) by 8.43 points, GPT-5.1 by 7.21 points, and baseline by 8.65 points—the largest margin in our evaluation, providing evidence that four-phase CoT with linguistic verification is particularly effective for grammar tasks. On AraTrust (safety and cultural trustworthiness), we achieve 90.22\% (+6.73 over baseline, +2.10 over GPT-5.1), approaching leader Qwen72b-ar-lora (91.40\%) within 1.18 points. On AlGhafa (multi-ability evaluation), we reach 81.88\%, establishing SOTA by surpassing both category leader Llama-3.3-70B (80.36\%, +1.52) and Falcon-H1 (80.50\%, +1.38), exceeding GPT-5.1 by 7.66 points. On ALRAGE, which evaluates RAG, Arabic-DeepSeek-R1 matches baseline performance while clearly exceeding category leader Qwen3-32B (+5.84) and GPT-5.1 (+4.52).

Arabic-DeepSeek-R1 achieves the highest OALL average (80.18\%), exceeding D2IL-Arabic-Qwen2.5-72B (+4.32), GPT-5.1 (+2.31), Falcon-H1 (+5.28), Jais-family (+14.75), and baseline (+6.56). Our reasoning-focused sparse MoE with novel four-phase CoT and 80/20 data curation yields the strongest overall OALL average while achieving SOTA or near-SOTA on 5 of 7 benchmarks.

\section{Discussion}

The results highlight that parameter-efficient adaptation of a reasoning-focused open model with an Arabic-heavy training mixture can deliver strong improvements in Arabic performance without requiring full retraining. Gains on ArabicMMLU and ArbMMLU demonstrate that the proposed pipeline substantially enhances multi-task and cross-lingual reasoning, closing most of the gap to a proprietary frontier model while exceeding the current OALL leaders on the majority MMLU-style benchmarks. This suggests that much of the remaining performance deficit for Arabic does not stem from architectural limitations, but from insufficient language- and culture-specific adaptation of otherwise strong base models. Because DeepSeek-R1 is a sparse MoE model \cite{fireworks2025}, only a subset of experts is active for any given token, which allows us to leverage a relatively large base model without incurring proportional training or inference cost \cite{zhang2025}. Sparse MoE designs have been shown to enable cost-effective training and deployment of very large language models by activating only a few experts per token while maintaining competitive quality. This property makes our Arabic adaptation computationally feasible while still benefiting from the capacity of a large-scale backbone. Crucially, this combination of sparse MoE architecture with parameter-efficient adaptation yields the highest OALL average (80.18\%) among all systems—surpassing D2IL-Arabic-Qwen2.5-72B (+4.32), GPT-5.1 (+2.31), Falcon-H1 (+5.28), and Jais-family (+14.75). Notably, Arabic-DeepSeek-R1 outperforms Falcon-H1-Arabic- 34B-Instruct on all seven benchmarks despite Falcon-H1 surpassing the ArbMMLU-HT and AlGhafa category leaders—demonstrating that adaptation of an open reasoning backbone can exceed even large dedicated Arabic systems that themselves beat specialized category leaders. To our knowledge, Arabic-DeepSeek-R1 is the first model to simultaneously achieve (a) the highest OALL average, (b) SOTA on a majority of benchmarks (5 of 7), and (c) dominant performance on grammar-focused tasks (+8.43 over category leader on MadinahQA), indicating that our sparse MoE adaptation, culturally grounded CoT, and strategic data curation together enable breakthrough Arabic performance across multiple dimensions. Leveraging this MoE-based base also removes the need to train an Arabic model from scratch, which would be prohibitively expensive in compute and environmental footprint for most academic or regional institutions \cite{guo2025}.

The dominant performance on MadinahQA and AraTrust indicate that the culturally grounded four-phase CoT supervision is particularly effective for tasks that demand precise language control and alignment with regional norms. In particular, the combination of a high-quality, contamination-controlled Arabic corpus \cite{bari2024} with targeted instruction and CoT supervision \cite{wei2022} appears critical not only to surpassing GPT‑5.1 on several benchmarks including MadinahQA and AraTrust (despite the proprietary model's larger undisclosed training budget), but also to pushing an open model to best-in-class average performance across the OALL suite.

Results on Arabic EXAMS and ALRAGE reveal some limitations. On EXAMS, category leader Llama-3.3-70B (66.67\%) remains ahead, suggesting exam-style questions may require curriculum-specific supervision. On ALRAGE, Arabic-DeepSeek-R1 matches baseline performance (+0.16) while substantially exceeding the category leader (+5.84) and GPT-5.1 (+4.52), but shows smaller relative gains than on grammar/safety benchmarks. This reflects our prioritization of reasoning over retrieval: CoT supervision targeted structured problem-solving rather than RAG workflows. Integrating retrieval-aware objectives in future work could address this gap.

Notably, our model achieves superior or at least competitive performance to GPT-5.1 on 5 of 7 benchmarks (MadinahQA, AraTrust, ALRAGE, AlGhafa, and Arabic EXAMS) as well as on the average across the entire OALL benchmark suite, demonstrating that open-source adaptation can match or surpass proprietary systems on most Arabic-specific evaluation dimensions.

Our results point to several directions for future work. First, integrating retrieval-aware training objectives or including RAG-style supervision during fine-tuning may help recover and surpass baseline performance on ALRAGE. Second, lightweight domain-adaptive phases targeted at exam curricula and RAG corpora could complement the current general-purpose supervision without substantially increasing computational cost. Third, further analysis of error patterns across dialects, domains, and safety categories could guide more fine-grained data curation and CoT design, strengthening both robustness and cultural alignment for Arabic users.

\section{Conclusion}

Arabic-DeepSeek-R1 establishes new SOTA on OALL (80.18\% average, +4.32 over open-source leader, +2.31 over GPT-5.1, +5.28 over Falcon-H1, +14.75 over Jais-family) through: (1) first application of reasoning- focused sparse MoE to Arabic via parameter-efficient adaptation, (2) novel four-phase, culturally-grounded CoT with explicit linguistic verification, and (3) contamination- controlled 80/20 Arabic-English data mixture. Our model achieves SOTA or near-SOTA on 5 of 7 benchmarks, with dominant grammar performance (MadinahQA: +8.43 over category leader), safety, multi-task understanding, and retrieval-augmented tasks, while revealing residual gaps on exam-style questions. Together, these findings suggest that much of the performance deficit for Arabic stems from under-specialization rather than inherent architectural limits, and that parameter-efficient adaptation of open MoE backbones offers a viable path to competitive, locally controlled Arabic LLMs without the cost of full pretraining.

\section*{Impact Statement}

Arabic-DeepSeek-R1 promotes digital equity by providing a sovereign alternative to Anglocentric proprietary systems, enabling local control and reducing dependence on external infrastructure for critical sectors. By demonstrating that an open, sparse MoE base model can be efficiently adapted—using an Arabic-specific training and evaluation recipe—to deliver SOTA on several key Arabic benchmarks and in average performance across the OALL suite, this work provides a concrete template for resource-constrained institutions to build high-quality, language- and culture-specific LLMs without industrial-scale compute. The resulting model has the potential to support more inclusive access to AI in Arabic-speaking regions across education, healthcare, public services, and safety-critical applications, where robust grammar, cultural alignment, and trustworthy behavior are essential.


\bibliography{example_paper}
\bibliographystyle{icml2026}

\newpage
\appendix
\onecolumn


\end{document}